%% file: main.tex
\documentclass[10pt,twocolumn,letterpaper]{article}

\usepackage[accsupp]{axessibility}  % Improves PDF readability for those with disabilities.

\usepackage{iccv}
\usepackage{times}
\usepackage{epsfig}
\usepackage{graphicx}
\usepackage{amsmath}
\usepackage{amssymb}

\input{macro}

\usepackage[pagebackref=true,breaklinks=true,letterpaper=true,colorlinks,bookmarks=false]{hyperref}

\iccvfinalcopy % *** Uncomment this line for the final submission

\def\iccvPaperID{6284} % *** Enter the ICCV Paper ID here

\ificcvfinal\fi

\usepackage{xcolor}
\usepackage{ifthen}
\usepackage{textcomp}
\definecolor{red}{rgb}{1,0,0}
\definecolor{slateblue}{rgb}{0.7,0.35,0.9}
\definecolor{green}{rgb}{0,0.5,0}
\definecolor{mahogany}{rgb}{0.75, 0.25, 0.0}
\definecolor{purple}{rgb}{0.6, 0, 0.6}
\definecolor{darkpurple}{rgb}{0.3, 0, 0.3}
\definecolor{darkgreen}{rgb}{0, 0.4, 0}
\definecolor{frenchblue}{rgb}{0.0, 0.45, 0.73}
\definecolor{blue}{rgb}{0,0,1}
\definecolor{goldenrod}{rgb}{0.65, 0.45, 0.03}
\definecolor{gray}{rgb}{0.5,0.5,0.5}
\definecolor{gold}{rgb}{1.0, 0.874, 0}
\definecolor{silver}{rgb}{0.67,0.67,0.67}
\definecolor{brown}{rgb}{0.8, 0.678, 0.4}

\begin{document}

%%%%%%%%% TITLE
\title{Hashing Neural Video Decomposition with Multiplicative Residuals in Space-Time}

\author{Cheng-Hung Chan \hspace{1.6em} Cheng-Yang Yuan \hspace{1.6em} Cheng Sun \hspace{1.6em} Hwann-Tzong Chen\\
National Tsing Hua University, Taiwan\\
% {\tt\small \{pierre5340, yuanemc0930692766\}@gmail.com} \hspace{1em} {\tt\small \{chengsun@gapp, htchen@cs\}.nthu.edu.tw}
}

%\twocolumn[{%
%\maketitle
%\renewcommand\twocolumn[1][]{#1}%
%    \centering
%    \includegraphics[width=.95\linewidth]{fig/top.pdf}
%    \vspace{-1.0em}
%    \captionof{figure}{
%    \textbf{Integration of layer-based video editing and spatiotemporal lighting rendering.}
 %   Our approach decomposes an input video into layers with multiplicative residuals that characterize complex spatiotemporal lighting variations.
%    Our method efficiently fuses the user-edited components with the expected lighting conditions to produce high-quality video output.
%    \label{teaser}\vspace{-0.5em}}
%    \vspace{7mm} 
%}]

\maketitle
\ificcvfinal\thispagestyle{empty}\fi

% TODO: 加一個放在頭的 figure (fig/top.pdf)，我不會加 QQ

%%%%%%%%% ABSTRACT
% pros of our work (TODO: 有一些還沒有補進去):
% - primary: 處理光影變化, secondary: 因此得到更乾淨的 object mask (前作會把光影變化 encode 在 object mask 當中)
% - 能判斷整個場景的光照並做到 camera 控制等等的操作
% - 支援高畫質 (480p 起跳) 與短 training 時間
% - inference speed: 12 fps，train 完後做完 post-processing 可以做到實時編輯

\begin{abstract}
We present a video decomposition method that facilitates layer-based editing of videos with spatiotemporally varying lighting and motion effects.
Our neural model decomposes an input video into multiple layered representations, each comprising a 2D texture map, a mask for the original video, and a multiplicative residual characterizing the spatiotemporal variations in lighting conditions.
A single edit on the texture maps can be propagated to the corresponding locations in the entire video frames while preserving other contents' consistencies.
Our method efficiently learns the layer-based neural representations of a 1080p video in 25s per frame via coordinate hashing and allows real-time rendering of the edited result at 71 fps on a single GPU.
Qualitatively, we run our method on various videos to show its effectiveness in generating high-quality editing effects. Quantitatively, we propose to adopt feature-tracking evaluation metrics for objectively assessing the consistency of video editing.
Project page: \url{https://lightbulb12294.github.io/hashing-nvd/}
\end{abstract}

%%%%%%%%% BODY TEXT

\begin{figure*}
\begin{center}
    \includegraphics[width=.95\linewidth]{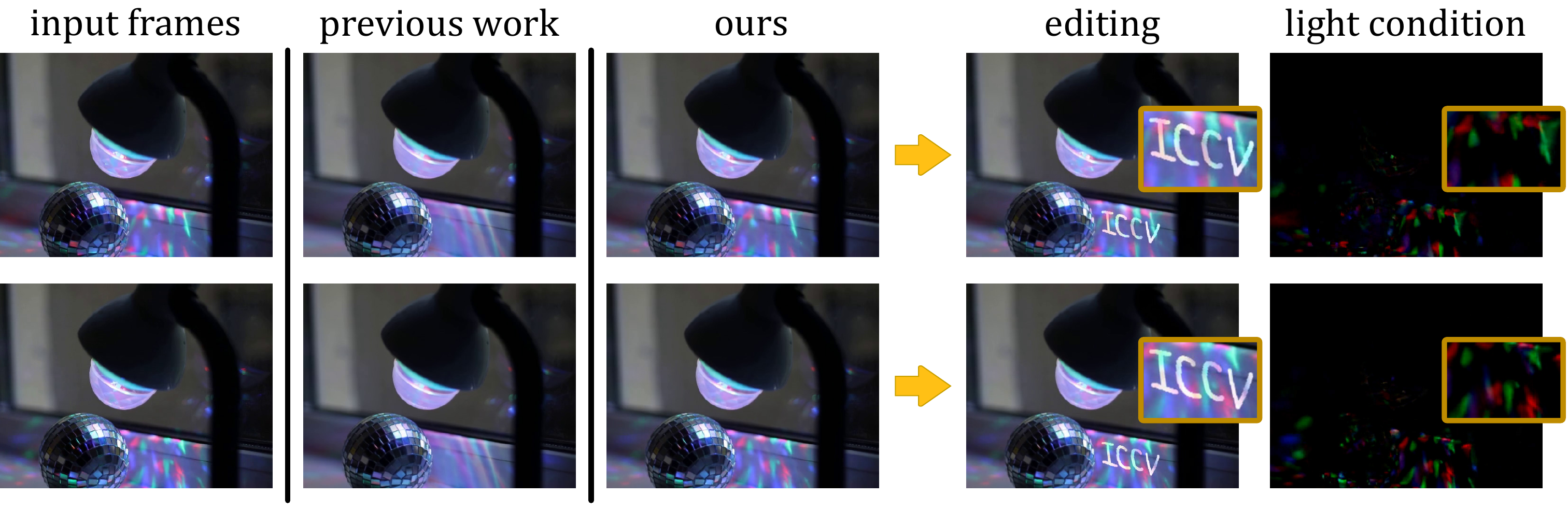}
\end{center}
   \vspace{-1.5em}
   \caption{
   \textbf{Integration of layer-based video editing and spatiotemporal lighting rendering.}
    Our approach decomposes an input video into layers with multiplicative residuals that characterize complex spatiotemporal lighting variations.
    Our method efficiently fuses the user-edited components with the expected lighting conditions to produce high-quality video output.
   }
\label{teaser}
\end{figure*}

\begin{figure*}
\begin{center}
    \includegraphics[width=0.980\linewidth]{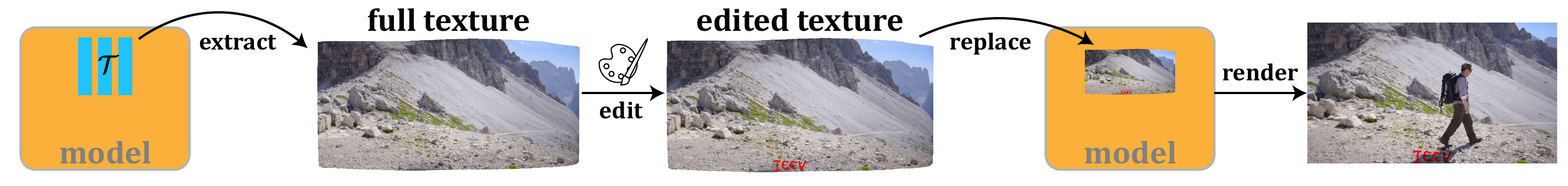}
\end{center}
   \vspace{-1.5em}
   \caption{
   \textbf{The procedure of layer-based video editing.}
   Our model allows users to apply edits to the extracted texture for rendering the edited video.
   }
\label{fig:edit_demo}
\end{figure*}

\section{Introduction}

% 從 image editing 帶入，講 video editing 的難處
%Modern image editing techniques have made it possible for even novice users to carry out modifications on complex scenes. 
Unlike image editing, video editing involves modeling the frame-to-frame relationships and addressing temporal variations such as motion and illumination changes. The task of video editing becomes challenging with the added dimension of time for a user who attempts to apply edits to a video while ensuring consistency across all frames. It is more convenient if we can handle the spatiotemporally varying components and reduce video editing to image editing---The user thus only needs to do edits on images with ease, and the editing results can seamlessly propagate to the entire video.

To achieve this goal, we may consider incorporating effective representations that can model and reconstruct the static and dynamic information in videos. Furthermore, for practical concerns, the algorithm must be efficient enough in modeling and rendering to support interactive editing. Recent work on video decomposition has proposed to employ neural-based representations, such as layered neural atlases \cite{KastenOWD21} and deformable sprites \cite{YeLTKS22}, to enable the conversion between the space-time video domain and 2D texture domain for editing. Despite their successes in showcasing impressive video editing effects, we notice that they often require longer training time or restrict to limited frame resolution. For example, it takes more than ten hours to derive the layered neural atlases \cite{KastenOWD21} from a 100-frame 480p video. Deformable sprites \cite{YeLTKS22} are relatively fast to derive but require more computation resources (over 24GB of GPU memory for a 480p video) and thus are unsuitable for editing high-resolution videos. We address the computation issue by incorporating hash grid encoding \cite{MullerESK22} into our framework and achieve fast training and rendering for high-resolution videos. Moreover, we introduce the multiplicative residual estimator to model the lighting variations across video frames, which can improve the reconstruction quality and allow illumination-aware editing unachievable by prior work, as shown in Fig.~\ref{teaser}.

% TODO: 描述一下自古至今的 work 是怎麼處理的，以及到現今為止依然有什麼樣的問題存在
% 問題: GAN-based editing 修改的範圍侷限於提供的 prior
% 問題: layered neural atlases 支援高畫質 (480p 以上)，但 training 時間太長 (10 hrs)
% 問題: deformable sprites 可以有限度的 non-rigid 的物體，速度也很快，但能 train 的 resolution 太低 (240p 時會花費 12G GPU RAM，480p 會超過 24G GPU RAM)，training 時間也會隨著 resolution 提升而指數成長

% TODO: 帶出我們的 work，整體而言相較之前有什麼樣的好處
% - primary: 處理光影變化, secondary: 因此得到更乾淨的 object mask (前作會把光影變化 encode 在 object mask 當中)
% - 能判斷整個場景的光照並做到 camera 控制等等的操作
% - 支援高畫質 (480p 起跳) 與短 training 時間
% - inference speed: 12 fps，train 完後做完 post-processing 可以做到實時編輯

% TODO: summary our contribution
% 我們是第一個可以輕鬆修改影片，且光照效果能漂亮的與做出的修改結合，而不需要任何對光源的 supervision
% training 速度很快，要求的資源很少，也可以 train 在高畫質或較長的影片上
% 我們演示了不同情境的影片修改，包含物體移動、光照變化、物件遮擋的修改等等 (camera motion control)

We summarize the contributions of this work as follows:
\begin{enumerate}
[topsep=0pt,partopsep=0pt,itemsep=4pt,parsep=0pt]
\item This work is the first to consider spatiotemporally varying lighting effects for layered-based video editing. The proposed multiplicative-residual estimator effectively decomposes the lighting conditions from the video without supervision. Our method can produce better-quality videos by fusing the edits with expected illuminations.
\item Our approach is efficient in both training and rendering. Compared with prior work, the proposed method improves the training time with fewer resources and thus enables training on higher-resolution or longer videos. The trained model can achieve fast video rendering via hashing-based coordinate inference. It takes about 40 minutes to train with a 1080p video of 100 frames on a single 3090 Ti GPU. The inference speed for rendering an edited video is 71 fps for 1080p resolution, allowing real-time video editing. 
\item The experimental results demonstrate appealing video edits in various challenging contexts, such as modifying the texture of moving objects, handling occlusion, and manipulating camera motion, where all can be fused with vivid lighting and shading for better effects that are not easy to achieve by prior work.
\end{enumerate}

\section{Related Work}
% TODO: more categories & more possible citing papers

% 相關性: 我們的 model 使用 video decomposition 來更好的解構 video，並提供較方便直覺的 editing
\paragraph{Video decomposition.}

Decomposing a video into layered representations is a long-standing video analysis approach in computer vision since the seminal work by Wang \& Adelson \cite{WangA94}.
Similar ideas have been continually revisited under different contexts with the development of new techniques.  
A typical formulation is motion segmentation that decomposes the video by motion~\cite{BrostowE99,JojicF01,KumarTZ08,WillsAB03}.
Video matting, on the other hand, focuses more on separating the alpha matte of the foreground object from the background for blending~\cite{Gu2022FactorMatteRV,LuCDZFR21,SenguptaJCSK20,XuPCH17}.
Recently, neural-based methods have shown to be effective in estimating layered representations for video segmentation and video editing~\cite{AlayracCAZ19,JampaniGG17,KastenOWD21,YangLS21,YeLTKS22}.
In this work, we also adopt neural networks to derive layered representations from videos. Further, inspired by the pioneering work in visual tracking for handling illumination changes~\cite{HagerB98}, we incorporate a new multiplicative residual representation to model the lighting variations for illumination-aware video editing.

% 相關性: 我們有做這件事
\paragraph{Video editing.}
Layered representations can benefit video editing in various ways. For example, the layered representations can serve as a visual proxy for intuitive video editing~\cite{KastenOWD21} or can be used to create the retiming effects~\cite{LuCDXZSFR20}. 
Editing can be more easily done on a unified texture map built from the video's background, such as background mosaics~\cite{CorreaM10}, tapestries~\cite{BarnesGSF10}, and layered neural atlases \cite{KastenOWD21}. Building a unified texture map for the foreground object is also useful, \eg, flexible sprites~\cite{JojicF01}, unwrap mosaics~\cite{Rav-AchaKRF08}, and deformable sprites~\cite{YeLTKS22}.
Another type of decomposition is to derive temporally-consistent intrinsic components from videos~\cite{Pfister14,Lin2017LayerBuilderLD} so that the edits can be performed on the reflectance for recoloring or texture transfer. Recent deep-learning-based methods mainly address a single task of video editing, \eg, video style transfer~\cite{WangYXL20,LiLK019,LaiHWSYY18,HuangWLMJZLL17,RuderDB16}.
or category-specific GAN-based video editing~\cite{AlalufPWZSLC22,PeeblesZ00ES22,TzabanMGBC22,XuAH22}.
Regarding achieving consistent video editing, it is crucial to know the temporal correspondences~\cite{JabriOE20, BianJEO22} so typical animation techniques like Rotoscoping~\cite{AgarwalaHSS04} can be applied.

%Text-guided video editing~\cite{Bar-TalOFKD22}.
%Text-based facial video editing~\cite{FriedTZFSGGJTA19}.

\section{Approach}

We formulate the problem of consistent video decomposition and editing as follows: The goal is to decompose an input video into multiple layers, where one layer comprises the background, and each of the remaining layers represents a foreground object.
We aim to facilitate user-friendly and intuitive editing of video components on time-independent static texture maps while ensuring the quality and consistency of the edited results in the output video. Fig.~\ref{fig:edit_demo} illustrates an editing procedure.
The model may rely on additional information obtained during pre-processing, such as backward and forward optical flow and rough object masks, to improve the training efficiency and consistency.

\subsection{Layer-Based Video Decomposition}
% 把影片拆解成多個 layer，並說明每個 layer 裡面包含的內容
A video of a scene typically contains multiple objects with distinct motions, appearances, shapes, and other attributes. Achieving consistent editing on the video scene and objects can be challenging. An efficient way to group objects is by decomposing the video into different motion layers. Static objects can be considered part of the scene's background, while only a few foreground objects move differently. Editing the objects on decomposed layers becomes more intuitive and straightforward since we can modify each object separately without touching the others.
We decompose videos into layers using backward and forward optical flow and coarse object masks, resulting in $N+1$ layers where layer $N$ represents the background, and the rest is for foreground objects.
Each layer $n$ comprises a predicted mask $\alpha_n$ across different frames, a time-independent texture map $M_n$, and the multiplicative residual $\ell_{n,t}$ characterizing the illumination for each frame $t$ built on the texture map.
Fig.~\ref{fig:overview} shows an overview of our model.

\paragraph{Layer hierarchy.} \label{sec:layer_hie}
% alpha 是每個 layer 最終的 mask，a 是 model predict 出來的 raw output
% 最後一個 layer (alpha_N) 是直接用最後剩下來的數字 (相當於固定 a_N=1)
% deformable sprites 也是用 hierarchy，需要在這裡 cite 他嗎?
Our alpha network $\mathcal{A}$ predicts the object mask probability for every layer.
% Rather than simply normalizing these probabilities and treating them as actual masks, we use them to create a front-to-back hierarchy.
Rather than simply normalizing the layer probabilities of each pixel and directly using them as layer masks, we build a front-to-back hierarchy.
That is, for layer $n$, its probability of object mask is computed by 
\begin{equation}
    \alpha_n=a_n\cdot\prod_{i=n+1}^{N} {(1-a_i)} \,,
\end{equation}
where $a\in[0,1]$ is the raw output of $\mathcal{A}$ with $a_N$ fixed and equal to $1$.
This hierarchy assumes that objects follow a strict front-to-back ordering, which also implies that layer $N$ represents the background.
Here, we specifically focus on scenes with only one foreground object. That is, $N=1$. We show an example of decomposing multiple foreground objects in the appendices.

\subsection{Texture Mapping}
% TODO: 講解 mapping network 和 texture network 的內容
We associate each object with its own texture, such that modifying a single pixel in the texture will impact the object's appearance in the rendered video scene.
We translate the spatiotemporal locations in the video into 2D UV coordinates for different layers by a mapping network $\mathcal{M}$. This allows us to sample colors on a texture with the translated coordinates.
We follow the approach of coordinate-based neural rendering \cite{KastenOWD21,MildenhallSTBRN20} and use the multi-layer perceptrons as our coordinate translator instead of a textured grid.

% TODO: 說明用 2D map 來 model objects / background (需要嗎?)
% TODO: 說明 texture 保持恆定 (不受時間影響)，作為該 object / background 的 ``true color'' 存在 (也許補充光照改變造成的顏色變化會在 residual network 處理)
% TODO: 說明不受時間影響的好處，可以把 texture grid sample 出來後進行修改，再透過 bilinear interpolation 插值出顏色來
%Textures are used to represent the appearance of objects and background in the scene. The texture is a 2D image that is mapped onto the surface of the object in the scene, which is fixed and remains constant throughout the video. Such design allows us to treat the texture as the ``true color'' of the object or background. 
%This means that any modification applied to the texture will be reflected consistently across all frames of the video, ensuring visual coherence.
Note that the appearance of an object in the scene may change throughout the video due to camera motion and illumination variations. We will explain how we address this issue with the proposed multiplicative residual in Sec.~\ref{sec:light_residual}.

The textures can be further obtained by grid-sampling the texture network $\mathcal{T}$. Once the texture is extracted, the texture network can be replaced by the extracted texture using UV coordinates from the mapping network, achieving the same effect via bilinear interpolation. This process enables users to modify and apply the extracted texture to the entire video.

\subsection{Multiplicative-Residual Estimator}\label{sec:light_residual}
% TODO: 提出我們的 residual network，可以處理光源/光照變化
% 使用 multiplicative 而非 additive 的原因
In order to apply a single modification to all video frames, the texture must remain constant throughout the entire video.
However, the color at the same location on an object may change over time due to variations in lighting conditions, resulting in low reconstruction quality or noisy textures. To address this issue and ensure high-quality reconstruction, we have developed a new method, {\em multiplicative-residual estimator}, to achieve constant texture across all frames while preserving illumination variations.

% TODO: 說明 residual network 的設計 (用 uvt 當 input)
% TODO: residual network 的意義
% TODO: 採用 multiplicative residual 的原因 (require sunset's help)
% - albedo rho.
%Our multiplicative-residual estimator takes a temporal UV-coordinate $(u,v,t)$ as input instead of $(x,y,t)$ since we aim to associate the lighting conditions with the appearance of the object itself rather than the video.
With this design, we can also synthesize illumination coefficients not present in the original video at specific time points, allowing us to manipulate camera motion.
Note that the prediction of the residual estimator is sharp. Therefore, we have designed losses to constrain the seen and unseen areas, as will be described in Sec.~\ref{sec:loss}.
The multiplicative-residual estimator predicts the illumination coefficient of each color channel on the texture map, which is multiplied by each channel to obtain the shaded color.

To provide more insight into our decision to use a multiplicative residual instead of an additive residual, we examine the process of diffuse shading:
\begin{equation}
    C = \rho L ~,
\end{equation}
where $\rho$ is the diffuse albedo term determined by surface material and $L$ is the lighting term determined by the environment.
When the lighting condition changes or a shadow is cast onto the surface, the new shaded color then becomes $C' = \rho \cdot L'$.
With multiplicative residual, we can reproduce $C'$ by scaling $C$ with $\frac{L'}{L}$.
In contrast, we have to add $C$ by $\rho (L' - L)$ if we use additive residual.
% Global illumination arises when some light $L(\omega_i)$ in direction $\omega_i$ is indirectly reflected from surface $\hat{n}$ in a scene.
% \begin{equation}
% C=\int_\Omega \frac{\rho}{\pi} L(\omega_i) (\omega_i\cdot \hat{n})d\omega_i \,,
% \end{equation}
% where $\rho/\pi$ is the diffuse term and $C$ is the reflected color. If $C'$ and $L'$ are the new shaded color due to variation of lighting condition, the coefficient of multiplicative residual $s$ would be
% \begin{equation}
% s=\frac{\int_\Omega L'(\omega_i)(\omega_i\cdot \hat{n})d\omega_i}{\int_\Omega L(\omega_i)(\omega_i\cdot \hat{n})d\omega_i}\,,
% \end{equation}
% and the coefficient of additive residual $b$ would be
% \begin{equation}
% b=\frac{\rho}{\pi}\left(\int_\Omega L'(\omega_i)(\omega_i\cdot \hat{n})d\omega_i-\int_\Omega L(\omega_i)(\omega_i\cdot \hat{n})d\omega_i\right)\,.
% \end{equation}
One can observe that the additive residual is more difficult to model as it also depends on the diffuse albedo term. We detail the practical result of the two choices in Sec.~\ref{sec:abla_residual}.

\begin{figure*}
\begin{center}
    \includegraphics[width=0.94\linewidth]{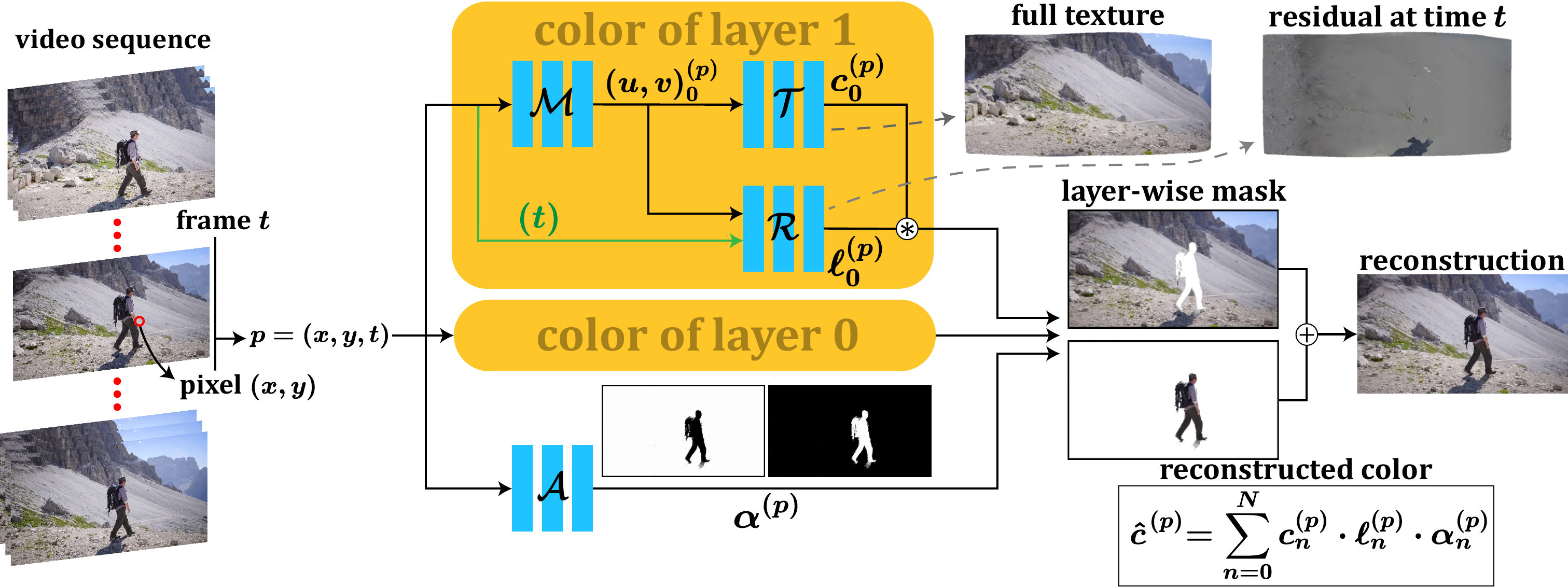}
\end{center}
\vspace{-1.5em}
   \caption{
   \textbf{Model pipeline.}
   Our model takes a video coordinate $p=(x,y,t)$ as input and decomposes the video into multiple layers.
   A representation for layer $n$ is modeled by a mapping network $\mathcal{M}$, a texture network $\mathcal{T}$, and a multiplicative-residual estimator $\mathcal{R}$; the three modules jointly convert a video coordinate $p$ into layered color $c_n^{(p)}$ and lighting coefficient $l_n^{(p)}$.
   An alpha network $\mathcal{A}$ also takes $p$ as input and predicts soft object masks $\boldsymbol{\alpha}^{(p)}$ for each layer.
   The final video color is reconstructed by the bottom-right equation. The multiplicative residuals are critical to our model for handling illumination variations.
   }
\label{fig:overview}
\end{figure*}

\subsection{Network Architecture} \label{sec:overview}
% 基本上和 layered neural atlases 大同小異，引用了 instant-ngp 的 hash grid encoding 來加快收斂速度
% 這整段我擔心會不會寫得跟 neural layered atlases 寫得太像? (model overview)
We design our model as an end-to-end framework trained via self-supervision. We take a spatiotemporal coordinate $p=(x,y,t)$ in the video as the model input and predict the reconstructed color of the corresponding pixel. Fig.~\ref{fig:overview} shows the pipeline of our model, which includes a mapping network, a texture network, and a multiplicative-residual estimator for each layered representation, with a shared alpha network generating the soft masks for different layers.

\paragraph{Mapping network.}
% UV coordinate (把 xyt 轉換到 texture 的座標上)
Following the design choice of recent coordinate-based approaches  \cite{KastenOWD21,MildenhallSTBRN20}, we build our network architecture with multi-layer perceptrons.
For each layer $n$, we pass a spatiotemporal coordinate $p$ through a mapping network to translate the video coordinate into a time-independent 2D texture-map coordinate $(u,v)$ by
\begin{equation}
    (u,v)_n^{(p)}=\mathcal{M}_n \left(p \right)\,, \quad (u,v)_n^{(p)}\in[-1,1]^2 \,,
\end{equation}
where $\mathcal{M}_n$ is the mapping network of layer $n$.

\paragraph{Texture network and multiplicative-residual estimator.}
% texture sample & residual sample (residual 吃 UV+t)
After obtaining the UV coordinates, the texture network takes the UV coordinates $(u,v)_n^{(p)}$ and produces the corresponding color at each position, while the multiplicative-residual estimator uses both the UV coordinates and the time $t$ as inputs to predict the corresponding illumination coefficients at each position: 
\begin{equation}
\begin{split}
    c_n^{(p)}&=\mathcal{T}_n\left((u,v)_n^{(p)}\right), \quad c_n^{(p)}\in[0,1]^3 \,, \\
    \ell_{n,t}^{(p)}&=\mathcal{R}_n\left((u,v)_n^{(p)}, t\right), \quad \ell_n^{(p)}\in\mathbb{R}_+^{3} \,,
\end{split}
\end{equation}
where $\mathcal{T}$ is the texture network and $\mathcal{R}$ is the multiplicative-residual estimator.
% Note that the full texture can be obtained by grid-sampling the texture network $\mathcal{T}$ and further modifying the full texture.

\paragraph{Alpha network.}
% alpha
We also predict an object mask for each layer to indicate the layer to which each pixel belongs:
\begin{equation}
    \boldsymbol{\alpha}^{(p)}=\mathcal{A}(p), \quad \boldsymbol{\alpha}^{(p)} \in[0,1]^{N+1} \,,
\end{equation}
where $\alpha_n^{(p)}$ is the probability that pixel $p$ belongs to layer $n$, \ie, $\sum_{n=0}^{N}{\alpha^{(p)}_n}=1$. The derivation of the object mask is detailed in Sec.~\ref{sec:layer_hie}.

\paragraph{Pixel color reconstruction.}
% combine them all
After obtaining the information described above, we can reconstruct the pixel color at position $p$ as 
\begin{equation}
    \hat{c}^{(p)}=\sum_{n=0}^{N} c_n^{(p)}\cdot \ell_n^{(p)}\cdot \alpha_n^{(p)} \,.
\end{equation}
 % 使用了 hash grid encoding
Inspired by \cite{MullerESK22}, we use hash grid encoding for all networks except for the mapping network, as texture coordinates are expected to be smooth. This approach is adopted to facilitate better convergence during training.

\subsection{Loss Terms} \label{sec:loss}
% TODO: optical flow (uv)、optical flow (alpha)、rgb + gradient、rigidity、alpha bootstrapping、sparsity、alpha regularization、residual regularization、residual consistent

% camera ready 新增
% from LNA: reconstruction loss, sparsity loss, optical flow loss, alpha bootstrapping loss
\paragraph{Losses inherited from previous work.}
We incorporate several losses from Layered Neural Atlases~\cite{KastenOWD21} to facilitate the training process.
These losses include 1) {\em optical flow loss}, which provides the model with the supervision of optical flow, and 2) {\em alpha bootstrapping loss}, which offers supervision of initial coarse masks; additionally, we use 3) {\em reconstruction loss} to stabilize the training and improve reconstruction quality, and 4) {\em sparsity loss} to avoid duplicate foreground objects in the texture area.
Further details can be found in the appendices.

\paragraph{Residual consistency loss.}
In order to produce significant yet smooth lighting conditions, the multiplicative residuals of different times $t_1$ and $t_2$ of the same position on the texture coordinate $(u,v)$, where $(u,v,t_1)$ corresponds to a visible area and $(u,v,t_2)$ corresponds to an invisible area, should be close in distribution and can have different intensities of illumination.
Formally, we have a small $k\times k$ patch $P$ at time $t_1$ on the video centered at $(x,y)$. We then sample their texture coordinates $(u,v)^{(P)}=\mathcal{M}(P)$.
The texture coordinates are then combined with different times $t_1$ and $t_2$ to get the corresponding multiplicative residuals:
\begin{equation}
\begin{split}
\ell^{(P)}&=\mathcal{R}\left((u,v)^{(P)}, t_1\right) \,,\\
\ell'^{(P)}&=\mathcal{R}\left( (u,v)^{(P)}, t_2 \right) \,.
\end{split}
\end{equation}
To this end, we encourage multiplicative residuals of the same position on texture coordinate should be positively correlated. Therefore, we define the residual consistency loss through normalized cross-correlation by 
\begin{equation}
\psi(P,t_2)=\frac{\left(\ell^{(P)}-\mu_{\ell^{(P)}}\right) \, \left(\ell'^{(P)}-\mu_{\ell'^{(P)}}\right) }{\sigma_{\ell^{(P)}} \, \sigma_{\ell'^{(P)}}} \,.
\end{equation}
We further introduce a variance-smoothness term that smooths the changes in lighting conditions as 
\begin{equation}
\mathbb{E}(P,t_2)=\sigma_{\ell'^{(P)}}^2 \,.
\end{equation}
We apply the above terms to all layers and get the overall loss as
\begin{equation}
\mathcal{L}_{\mathcal{R}\mathrm{con}}=\lambda_{\mathcal{R}\mathrm{con}}(\psi + \beta\mathbb{E}) \,,
\end{equation}
where we choose $k=3$ since our patch size is $3\times3$ and $\beta=16$.

\paragraph{Residual regularization.}

Minimizing the current losses can collapse to a trivial solution because the multiplicative-residual estimator absorbs all colors.
We regularize the residuals equal to $1$ since ``true color'' of an object in the video depends on the overall light conditions. That is, if the light source is a pure blue light in the entire video shining on a white wall, we can only say that the steady color of the wall is ``blue'', as we do not have enough information to determine the true color of the wall.
\begin{equation}
\mathcal{L}_{\mathcal{R} \mathrm{reg}}=\lambda_{\mathcal{R}\mathrm{reg}}||\mathcal{R}(\cdot)-1||^2_2 \,.
\end{equation}

\paragraph{Alpha regularization.}
We introduce an additional regularization term to address potential issues with the alpha network. Specifically, the sparsity loss used in the alpha network tends to assign a value of zero (\ie, black) to regions that are not visible in the input video, which can result in noisy masks and incorrect shadow embedding. To mitigate this problem, we enforced a constraint that each pixel should contribute to at most one layer. This ensures that the mask for each layer is clean and reliable and that the lighting conditions are properly embedded in each layer.
\begin{equation}
\mathcal{L}_{\alpha\mathrm{reg}}=\lambda_{\alpha\mathrm{reg}}\text{BCE}\left( \max_{n\in \{0,\cdots,N\}}\alpha_n \right) \,,
\end{equation}
where $\text{BCE}$ is binary cross entropy.

\subsection{Hash Grid Encoding}
Like \cite{KastenOWD21, MildenhallSTBRN20}, we have considered employing positional encoding as the input encoding for the multi-layer perceptrons. The purpose of including the positional encoding is to enrich the discriminative power of the coordinates and ensure that high-frequency details can be adequately represented.
Inspired by~\cite{MullerESK22}, we choose to adopt hash grid encoding as the input encoding method for our model. By leveraging this technique, the input features are encoded into a set of intermediate representations that span a broad spectrum of spatiotemporal resolutions, ranging from coarse- to fine-grained, and can flexibly adjust to distinct regions while maintaining high levels of accuracy and consistency.
Formally, the 2D or 3D input feature is treated as a coordinate of multi-resolution grids and then used to sample data from the grids using interpolation techniques. The resulting sampled data are concatenated and passed into multi-layer perceptrons to obtain the final output.

%% Implementation details are kept in temp_for_sup.tex

\section{Experiments}
% TODO
We conduct our experiments on the DAVIS dataset~\cite{PerazziPMGGS16} and various internet videos to demonstrate the effectiveness of our approach in video reconstruction and consistent video editing. We also design an evaluation metric of \emph{edit consistency} on the TAP-Vid-DAVIS dataset~\cite{doersch2022tapvid}.
In addition, we use our method to generate videos with different camera motions from those in the original input video.
Furthermore, we conduct ablation studies on the multiplicative-residual estimator and encoding type to assess the impact on the performance of different architecture choices.

\begin{figure*}
\begin{center}
    \includegraphics[width=0.97\linewidth]{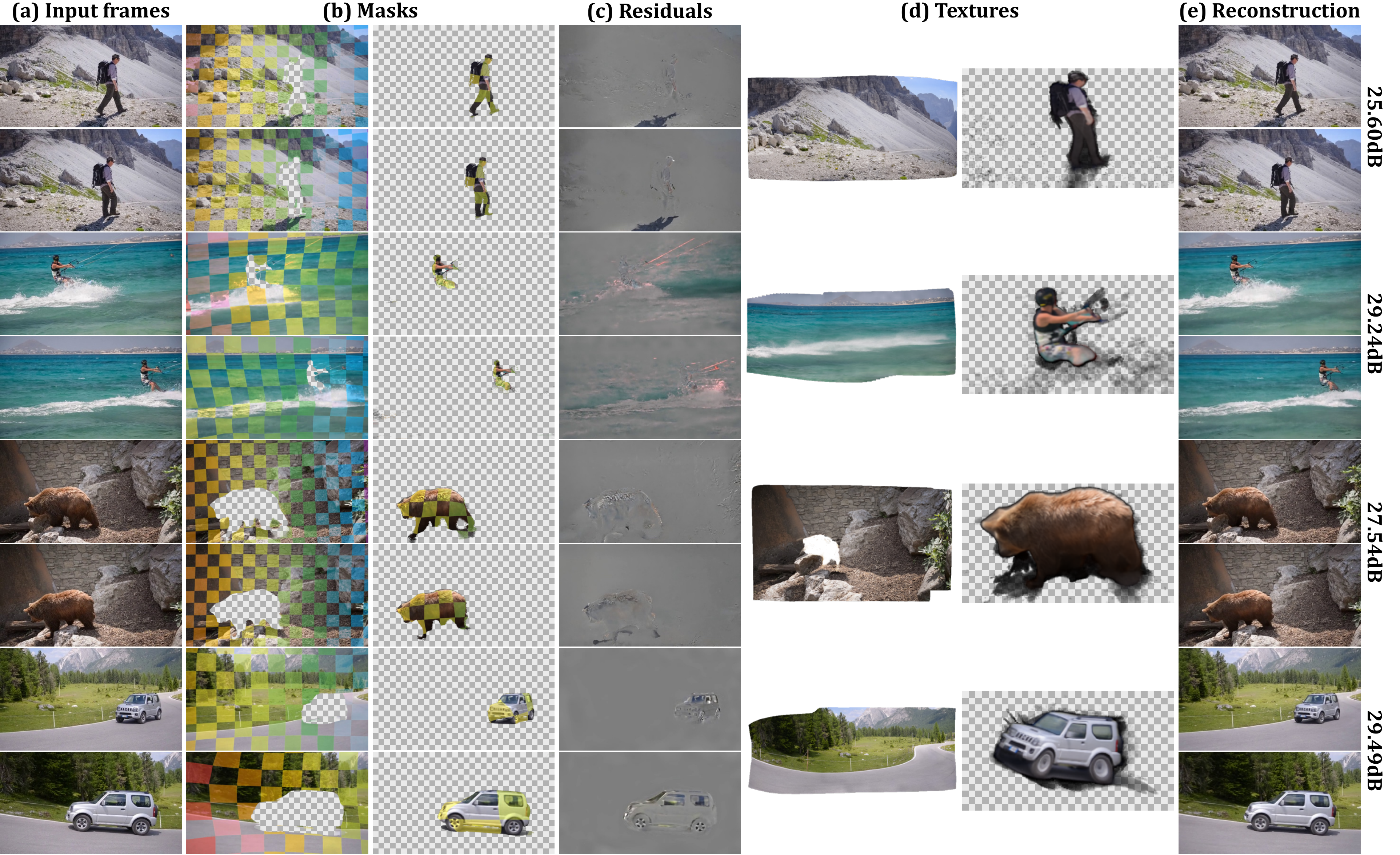}
\end{center}
\vspace{-1.5em}
   \caption{
   \textbf{Qualitative results on the DAVIS dataset.}
   We show our predicted masks, textures, m-residuals, reconstructed results, and PSNR of different videos. We overlay a color checkerboard on the masked areas to visualize the texture transformations.
   }
\label{fig:qualitative}
\end{figure*}

\subsection{Qualitative Results} \label{sec:qualitative}
% 放上我們 work 的結果
% blackswan(O), bear(O), hike(O), car-turn(O), kite-surf(O), libby(O)
% LNA 放相同解析度的，deformable sprites 只能放次等解析度，順便解釋原因 (耗費 GPU 資源過多，強調連在相同層級下 train 都有困難)

Fig.~\ref{fig:qualitative} shows four qualitative results of our work on the DAVIS dataset~\cite{PerazziPMGGS16}. For each video frame, 
we show the predicted masks (second and third columns), the m-residuals (fourth column), the corresponding layered textures (fifth and sixth columns), and the reconstruction result (last column). Please also refer to
Fig.~\ref{teaser} for an example result highlighting the robustness of our m-residual estimation. More results and higher resolution reconstruction can be found in the appendices.

With merely the guidance from the coarse object mask provided by Mask R-CNN~\cite{HeGDG20}, our model successfully separates the object from the background yielding a clean and precise mask.
In the second video in Fig.~\ref{fig:qualitative}, we showcase an example of a video occluded by complex water splashes. Our model can accurately reconstruct the video without introducing noticeable texture distortions.
We superimpose a rainbow checkerboard on the mask to visualize the transformation of each layer. It is evident that each coordinate's component remains at the same position in the texture.

Since the multiplicative residual may brighten or darken the texture, we display the m-residual output on a gray canvas to visualize its effect.
In the first video, our m-residual estimator successfully models the shadow of the hiker while the background texture retains its original color.
The m-residual estimator takes both the texture coordinate and time as the input, allowing it to learn how to model the changes in lighting conditions through space-time and adjust the color accordingly. Such a mechanism enables our model to generate realistic and consistent shading effects for different objects in the video, such as the shading on the bear's fur or the reflection on the hiker's hat.
We also report the PSNR of each  video as evidence of high-quality reconstruction.

\begin{figure*}
\begin{center}
    \includegraphics[width=0.97\linewidth]{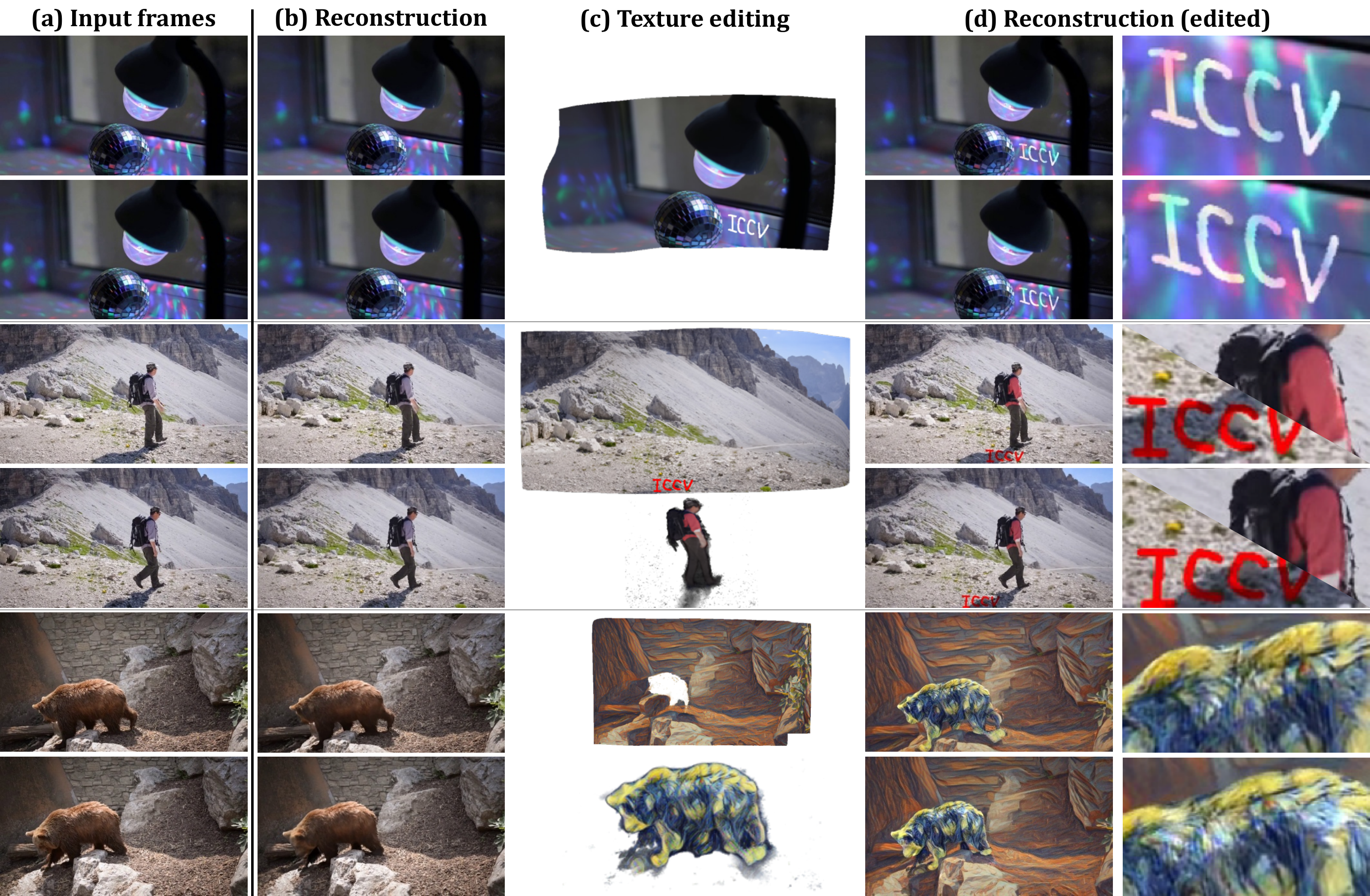}
\end{center}
\vspace{-1.5em}
   \caption{
   \textbf{Edits on videos with varying lighting conditions.}
   We apply various editing techniques on different components of the videos to observe the corresponding changes in the lighting conditions. The textures with edits are shown in (c), while the editing results are enlarged in the right column of (d). The lighting and shading fuse with the edits flawlessly.
   }
\label{fig:qual_light}
\end{figure*}

% camera ready 新增
\subsection{Comparison with Previous Work}

We report the PSNR values and various settings for every approach in Table~\ref{tab:comparison}.
We also provide more quantitative comparisons with Deformable Sprites (DS)~\cite{YeLTKS22} and Layered Neural Atlases (LNA)~\cite{KastenOWD21} under multiple metrics in Table~\ref{tab:more_comparison}.
Our method demonstrates superior performance of PSNR, rendering speed, and GPU memory consumption compared to previous methods on all three videos while maintaining the same or better resolution.
Note that we could not run Deformable Sprites for the $768\times 432$ resolution, as Deformable Sprites failed to finish due to their GPU memory usage exceeding the limit of a single GPU with 24 GB memory.
In contrast, our method can efficiently decompose higher-resolution videos into full-HD layers using affordable computation resources while maintaining comparable reconstruction results.
To compute the PSNR, we scale the input video up to the reconstructed resolution. Therefore, the PSNR of our 1080p model might be slightly lower than that of the $768\times432$ model, as the PSNR of the 1080p model is calculated at the corresponding resolution and considered harder due to having more details to reconstruct than the low-resolution counterpart.

\subsection{Consistent Video Editing}
% video editing (ICCV)
% style transfer?
% texture re-modeling (hike 衣服的顏色)
For editing purposes, we use a $1000\times1000$ grid sampling to render the texture networks. We then modify the rendered texture and sample color via bilinear interpolation.
As our key advantage lies in the m-residual representation of illumination, we have focused our editing on videos or components that exhibit changes in lighting.

We show our editing results in Fig.~\ref{fig:qual_light}. We choose three videos to present the rendering quality. The first video is sourced from internet videos, while the other two are taken from the DAVIS dataset~\cite{PerazziPMGGS16}.
For evaluating the editing quality, we modify the background texture of the first two videos by adding handwritten characters of ``ICCV''. We also adjust the color of the shirt of the hiker in the second video. Finally, we perform style transfer on the third video on both background and object textures.
In the first video, despite the complexity of the lighting conditions caused by the Disco ball, our approach can successfully handle and diffuse the light onto the edited region, resulting in high-quality output. Previous approaches fail to achieve such a representation of complex lighting conditions.
In the second video, the color of the shirt changes convincingly according to the different lighting conditions in two different frames, while our edit on the background is successfully occluded by the shadow.
We can see that in the third video, our edited reconstruction follows the textures and the modified lighting conditions fit seamlessly with our modifications, as shown in (d) of the fifth and sixth rows.
In the appendices, we also provide additional editing results of various types to further demonstrate the versatility and effectiveness of our method.

\begin{table*}
    \begin{center}
    \begin{tabular}{c|cccc|ccc}
    \small
    & & Training & Rendering & GPU & & PSNR & \\
    Method & Resolution &  time &  speed (fps) &  memory & {\sf bear} & {\sf disco ball} & {\sf hike} \\ \hline
    % PSNR 的計算都是把 input video scale，所以低解析度算出來的數值會比較好看一點
    Deformable Sprites~\cite{YeLTKS22}& $213\times120$ & 10 minutes & 5  & 5 GB & 22.7 & 26.2 & 21.5 \\
    Deformable Sprites~\cite{YeLTKS22} & $427\times240$ & 20 minutes & 1.6  & 12 GB & 23.6 & 26.4 & 22.0 \\
    Layered Neural Atlases~\cite{KastenOWD21} & $768\times432$ & 5.5 hours & 0.5  & 3 GB & 27.3 & 29.0 & 25.2 \\
    Ours & $768\times432$ & 40 minutes & 787  & 3 GB & 27.5 & 37.7 & 25.6 \\ \hline
    Ours & $1920\times 1080$ & 40 minutes & 71  & 5 GB & 26.9 & 35.6 & 25.4
    \end{tabular}
    \end{center}
    \vspace{-1.5em}
    \caption{
    \textbf{Comparison results.}
    We report the PSNR for each video, along with the corresponding training time, rendering speed (frames per second), and GPU memory usage under different resolutions.
    Our work achieves better results than prior work on the three videos, achieving faster rendering, lower GPU memory consumption, and higher resolution. Note that we measure the PSNR at the corresponding reconstructed resolution. Therefore the PSNR tends to favor the evaluation at low resolution, explaining why our 1080 reconstructions have slightly lower PSNR than our 480p reconstructions. 
    \label{tab:comparison}
    }
\end{table*}

\begin{table}[!h]
\vspace{-1.4em}
\scriptsize
\centering
    \begin{tabular}{@{}c@{\hskip 4pt}|@{\hskip 4pt}c@{\hskip 4pt}c@{\hskip 4pt}c@{\hskip 4pt}|@{\hskip 4pt}c@{\hskip 4pt}c@{\hskip 4pt}c@{\hskip 4pt}|@{\hskip 4pt}c@{\hskip 4pt}c@{\hskip 4pt}c@{}}

    & \multicolumn{3}{c@{\hskip 4pt}|@{\hskip 4pt}}{\sf bear} & \multicolumn{3}{c@{\hskip 4pt}|@{\hskip 4pt}}{\sf disco ball} & \multicolumn{3}{c@{\hskip 4pt}}{\sf hike} \\
    Method & PSNR & LPIPS & SSIM & PSNR & LPIPS & SSIM & PSNR & LPIPS & SSIM \\
    \hline
    % PSNR 的計算都是把 input video scale，所以低解析度算出來的數值會比較好看一點
    DS & 23.6 & 0.23 & 0.78 & 26.4 & 0.21 & 0.89 & 22.0 & 0.29 & 0.68 \\
    LNA & 27.3 & 0.19 & 0.85 & 29.0 & 0.11 & 0.95 & 25.2 & 0.19 & 0.80 \\
    Ours & \textbf{27.5} & \textbf{0.16} & \textbf{0.87} & \textbf{37.7} & \textbf{0.04} & \textbf{0.97} & \textbf{25.6} & \textbf{0.16} & \textbf{0.82} \\
    \hline

     & \multicolumn{3}{c@{\hskip 4pt}|@{\hskip 4pt}}{\sf kite-surf} & \multicolumn{3}{c@{\hskip 4pt}|@{\hskip 4pt}}{\sf car-turn} & \multicolumn{3}{c@{}}{\sf libby} \\
     % & PSNR & LPIPS & SSIM & PSNR & LPIPS & SSIM & PSNR & LPIPS & SSIM \\
     \hline

     DS & 21.2 & 0.34 & 0.62 & 22.2 & 0.32 & 0.65 & 21.6 & 0.41 & 0.57 \\
     LNA & 28.2 & 0.30 & 0.76 & 27.5 & 0.30 & 0.88 & 29.4 & 0.31 & 0.91 \\
     Ours & \textbf{29.2} & \textbf{0.26} & \textbf{0.78} & \textbf{29.5} & \textbf{0.23} & \textbf{0.91} & \textbf{29.6} & \textbf{0.26} & \textbf{0.92} \\

    \end{tabular}
    \vspace{-1em}
    \caption{
    \textbf{More comparison results.}
    Our model outperforms previous methods in all videos and achieves better scores in all evaluation metrics.
    % DS refers to Deformable Sprites, LNA stands for Layered Neural Atlases.
    % DS results are trained at 240p resolution, while both LNA and our method are trained at 480p resolution.
    % % kite-surf 在 main paper 上的數據填錯了: 不是 25.24，而是 29.24
    \label{tab:more_comparison}
    }
\end{table}

\begin{table}
    \vspace{-1em}
    \begin{center}
        % \begin{tabular}{cc|c|c|c|c}
        %      & & black swan & parkour & kite-surf & cows\\
        %     \hline
        %      & AJ & 0.81 & 0.04 & 0.48 & 0.51\\
        %     ours & $<\delta_{avg}^x$ & 0.87 & 0.10 & 0.57 & 0.61\\
        %      & OA & 1.00 & 0.48 & 0.97 & 0.99 \\
        %      & & black swan & parkour & kite-surf & cows\\
        %     \hline
        %      & AJ & 0.39 & 0.07 & 0.25 & 0.55\\
        %     RAFT~\cite{TeedD20} & $<\delta_{avg}^x$ & 0.52 & 0.17 & 0.35 & 0.65\\
        %      & OA & 1.00 & 0.48 & 0.86 & 0.99
        % \end{tabular}
        \begin{tabular}{@{ }c@{ }|c@{ }c@{ }c|c@{ }c@{ }c}
         & & ours & & & RAFT & \\
         & AJ & $<\delta_{avg}^x$ & OA & AJ & $<\delta_{avg}^x$ & OA \\
         \hline
         black swan & 0.81 & 0.87 & 1.00 & 0.39 & 0.52 & 1.00 \\
         %\hline
         parkour & 0.04 & 0.10 & 0.48 & 0.07 & 0.17 & 0.48 \\
         %\hline
         kite-surf & 0.48 & 0.58 & 0.99 & 0.25 & 0.35 & 0.86\\
         %\hline
         cows & 0.51 & 0.61 & 0.99 & 0.55 & 0.65 & 0.99
        \end{tabular}
    \end{center}
    \vspace{-1.6em}
        \caption{
    \label{tab:quan_tapvid}
        \textbf{Quantitative results on TAP-Vid.}
        We test the editing consistency on a subset of TAP-Vid DAVIS and show the result for each video. The reported metrics, the higher the better, are average Jaccard (AJ), average position accuracy ($<\delta_{avg}^x$), and binary occlusion accuracy (OA).
        Compared to RAFT~\cite{TeedD20}, our method performs better or reaches comparable results on all videos.
    }
\end{table}

\subsection{Quantitative Results for Editing Consistency}
% eval on tapvid (好做的跟難做的都要放: blackswan、libby、parkour)
Our work mainly focuses on consistent video editing. However, famous metrics such as PSNR (as we show in Fig.~\ref{fig:qualitative}) or other reconstruction quality measures do not necessarily reflect the quality of consistent video editing.
We take advantage of TAP-Vid~\cite{doersch2022tapvid}, which provides several evaluation metrics and the ground-truth feature tracking on the foreground object.
We use the feature points of the first unobstructed frame $(f_{0,x},f_{0,y})$ as our base point and convert it into texture coordinate $(u_0, v_0)$.
For all other frames $t$, we select the video coordinate whose mapped texture coordinate is closest to $(u_0,v_0)$ as our prediction for frame $t$. Therefore, the feature points on the same track should be mapped to the same coordinate on the foreground texture for an ideal rendering result.

We report our results in Table~\ref{tab:quan_tapvid}. We select a subset of the TAP-Vid DAVIS dataset for our quantitative analysis. For reference, we also report the results of RAFT~\cite{TeedD20}, as a baseline, on the same videos since our optical flow supervision comes from it. In general, the correspondences obtained by our method are more consistent than those by RAFT, as RAFT considers only neighboring frames while ours works on a unified foreground representation.
Specifically, our method achieves higher scores than RAFT on the ``black swan'' and ``kite-surf'' videos and comparable results on the other two videos. The low score on the ``parkour'' video is due to the complex geometric changes in the video, which our current architecture can reconstruct well but edit poorly on the foreground parkour runner. Nevertheless, the editing on the background still exhibits comparable quality.

\subsection{Manipulating Camera Motion}
% 固定視角
% 轉圈圈
% 固定視角把 hike 人放到 blackswan 池子裡
% cemera view 可以用多種方式 sample
% 只要把 camera view 定出來，顏色和光照都可以輕易取得
% 對於 object mask 來說，則是需要找到 UV map (object 與 background) 之間的對應關係，然後把 mask value interpolate 上去才行 (用 scipy LinearNDInterpolator 做到)。至於未知的資訊部分，則全部視作 background 看待
With our multiplicative residual estimator, we are able to synthesize realistic views that are not shown in the input video.
A different camera view can be synthesized by scaling, shifting, or rotating the original video coordinates $(x,y)$ to get $(x',y')$ and concatenating it with the temporal information $t$.
Once the new set of video coordinates $p'=(x',y',t)$ is obtained, we can use the mapping network $\mathcal{M}$ to obtain the texture coordinate $(u,v)^{(p')}$ for each layer. The color $c^{(p')}$ and the lighting $\ell^{(p')}$ of all layers can be obtained based on these texture coordinates.
For each non-background layer $n$, we use the alpha network $\mathcal{A}$ to obtain the original soft mask $\alpha^{(p)}$ for each $p=(x,y,t)$ and establish a relation as $(u,v)_0\to\alpha^{(p)}_n$ if $(u,v)_0=\mathcal{M}_0(p)$.
Then, we linearly interpolate the value of the mask at each location $p'$ by triangulating the input data.
That is, we establish a correspondence between the coordinates of the object textures and the background texture and then transfer the mask values from the object onto the background texture.

\begin{figure}
\vspace{-1.5em}
\begin{center}
    \includegraphics[width=1.0\linewidth]{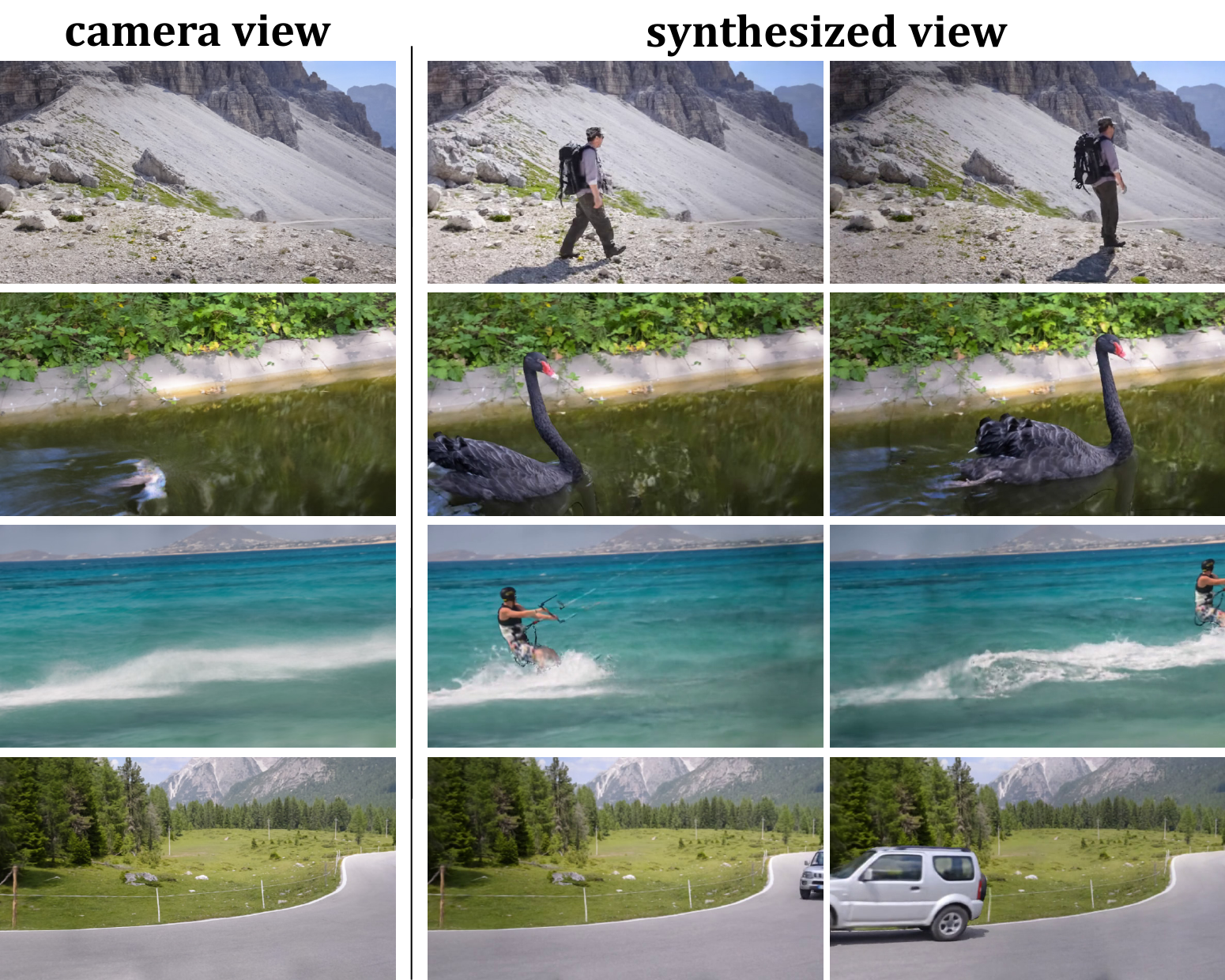}
\end{center}
\vspace{-1.5em}
   \caption{
   \textbf{Camera motion manipulation.}
   %We fix the camera at a specific position. 
   %Our multiplicative residuals affect the lighting outside the given video frame.
   The multiplicative residual serves as a smooth representation that effectively adjusts the lighting conditions with a fixed camera.
   }
\label{fig:camera_control}
\end{figure}

\begin{figure}
\begin{center}
    \includegraphics[width=1.0\linewidth]{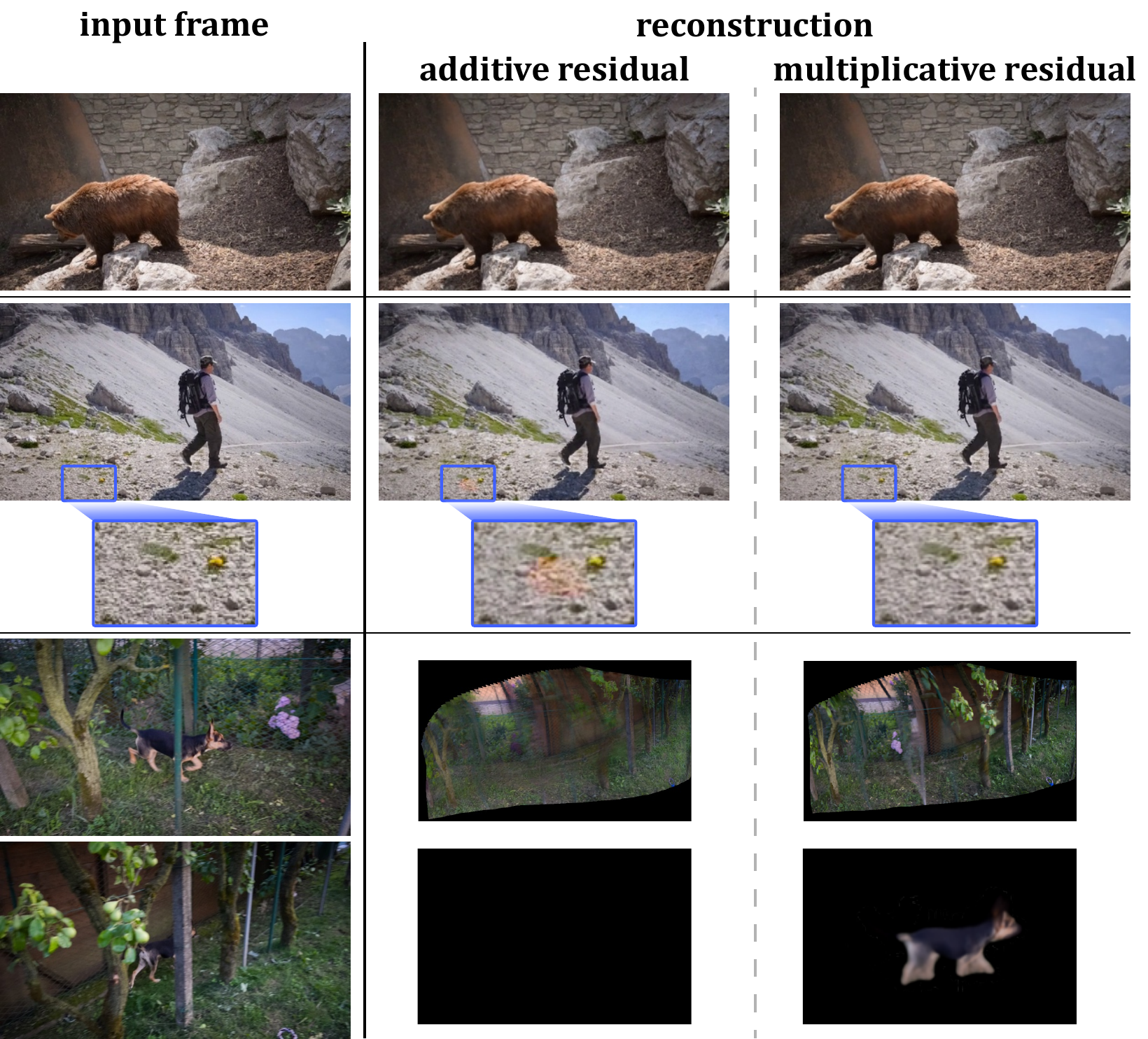}
\end{center}
\vspace{-1.5em}
   \caption{
   \textbf{Choice of the residual estimator architecture.}
   Using an additive residual estimator architecture may introduce artifacts or degrade the quality of textures, while a multiplicative one works correctly.
   }
\label{fig:ablation_add}
\end{figure}

% camera ready 新增
\begin{figure}
\vspace{-0.8em}
\begin{center}
    \includegraphics[width=1.0\linewidth]{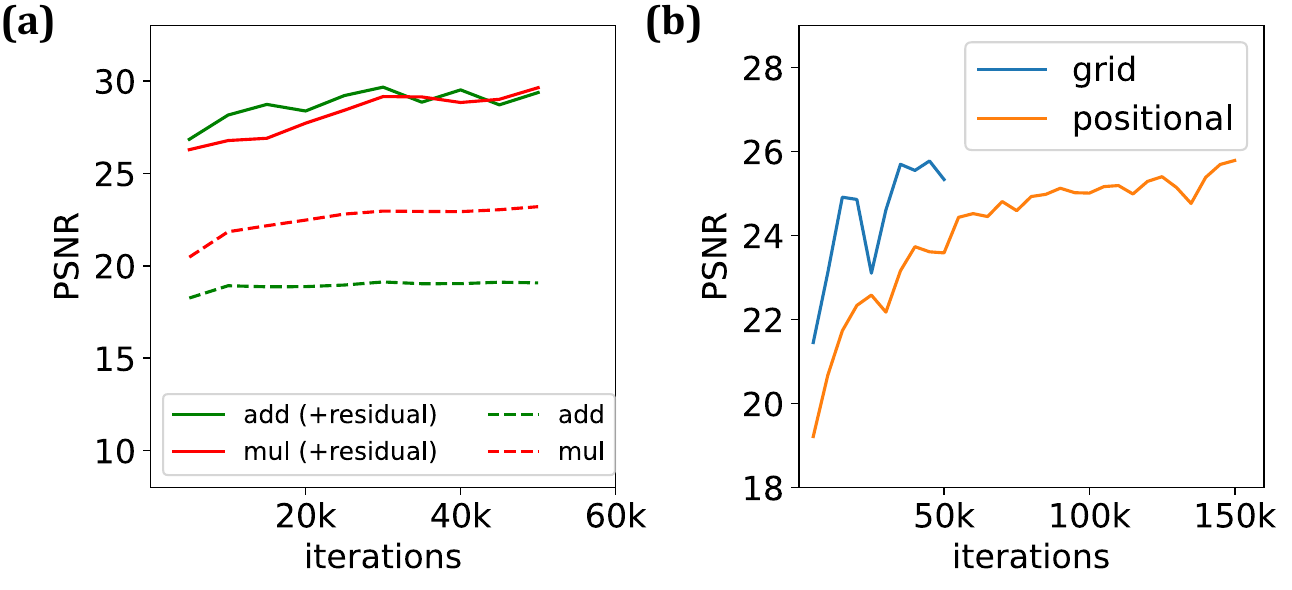}
\end{center}
\vspace{-1.7em}
    \caption{
    \textbf{Quantitative ablation studies.}
    (a) The reconstruction PSNR degradation of the additive residual is significantly higher than that of the multiplicative residual.
    (b) Hash grid encoding can achieve better PSNR in fewer iterations. % rebuttal 裡面寫說要加更多 iteration，但目前沒有資源
    \label{fig:ablation_quan}
    }
\end{figure}

We set up a virtual camera in the background to capture the entire video scene from a fixed position. Fig.~\ref{fig:camera_control} shows our result.
The camera is fixed (\ie, the background remains static) while the object keeps moving in the scene. In the fourth video, the car is partially visible in the beginning frames and fully visible later, with its side facing the camera.
The third video demonstrates that the splashes consistently follow the entire path of the original camera view. With our multiplicative-residual estimator, we can successfully reconstruct the surfing scene while accurately representing the water splashes.
We show more results with varying camera motions in the appendices.

\subsection{Ablations}
% residual (difference between * and +)
% residual consistent loss smooth term?

\paragraph{Residual type.}\label{sec:abla_residual}
We present an ablation study to verify the choice of the architecture of our residual estimator. As mentioned in Sec.~\ref{sec:light_residual}, we adopt the multiplicative residual instead of the additive residual. We show the comparisons of the two designs in Fig.~\ref{fig:ablation_add} and Fig.~\ref{fig:ablation_quan}a.
The additive residual estimator is conducted, and the output is normalized to $[-1,1]$ to ensure they have the same representation capability. In Fig.~\ref{fig:ablation_add}, both additive and multiplicative residuals provide promising reconstruction results for the first video. However, when it comes to the complex texture, like the background of the second video, the additive residual fails to represent it and compensates for the wrong illuminations. In the third video, the additive residual fails to represent precise background texture, and the dog vanishes from the texture.
Fig.~\ref{fig:ablation_quan}a demonstrates a significant drop in PSNR when ablating the residual module. This finding reflects that additive residual is prone to overfit the color that should be attributed to the editable texture.
In contrast, the multiplicative residual provides complete background and object textures.

\paragraph{Hash grid encoding.}
We evaluate the effectiveness of hash grid encoding in Fig.~\ref{fig:ablation_quan}b.
Positional encoding needs more MLP layers to achieve good quality, which takes 1.5$\times$ processing time per iteration compared to hash encoding. However, hash grid encoding still achieves better PSNR in fewer iterations.

% camera ready 新增
\section{Discussion}
We have observed several interesting points that are worth discussing.
% Can the residual model shadow casting to moving foreground?
% Some distortion on the edited parts. Imperfect bear alpha masks.
Our method relies on Mask-RCNN and RAFT to provide the foreground mask and optical flow as external priors.
As a result, our method may be biased by the inaccurate learning-based priors on some difficult cases like the spinning lighting in {\sf disco ball} or the object boundary in {\sf bear}.
% Object edge artifact.
The reconstructed objects in the video may exhibit edge artifacts due to referencing the incorrect texture layer.
% Smooth texture of bear.
% possible solutions
Our method with reconstruction loss could improve the initial external prior but is still not perfect to solve all the artifacts.
Adopting better external priors or designing strong internal priors are both good future improvements.
The problem of object edge artifacts is similar to the seam artifact in mesh uv-parameterization from which future exploration may take inspiration.

\section{Conclusion}
We have presented a neural layer decomposition method to facilitate illumination-aware video editing. The proposed multiplicative-residual estimator can effectively derive the layered representation that characterizes the spatiotemporally varying lighting effects. We have also implemented hash grid encoding for fast coordinate inference. Our model, therefore, significantly reduces the training time and achieves real-time rendering speed with a low requirement of computation resources, enabling interactive editing on high-resolution videos. We use our model to generate high-quality video editing results, where, in particular, the varying illumination effect can only be achieved by ours rather than the previous methods.

% camera ready 新增
\section*{Acknowledgments}
This work was supported in part by NSTC grants 111-2221-E-001-011-MY2 and 112-2221-E-A49-100-MY3 of Taiwan. We are grateful to National Center for High-performance Computing for providing computational resources and facilities.

\section*{Appendices}

\section*{A. More Experimental Results}

This section summarizes our supplementary results with links to the corresponding videos on our project website at \url{https://lightbulb12294.github.io/hashing-nvd/}.

\subsection*{A.1. Video Reconstruction}

We present additional reconstruction results on the DAVIS dataset~\cite{PerazziPMGGS16}, specifically, {\sf bear}, {\sf black swan}, {\sf car-turn}, {\sf cows}, {\sf hike}, {\sf kite-surf}, {\sf libby}, {\sf lucia}, and {\sf parkour}, as well as internet videos such as {\sf disco ball}~\cite{VideoDiscoBall} and {\sf sunset}~\cite{VideoSunset}.
These results are compiled into a single video and are showcased in the \href{https://lightbulb12294.github.io/hashing-nvd/#Reconstruction}{\sf Reconstruction} section of our project page.
We also provide several full-resolution (1080p) video reconstructions, including {\sf bear}, {\sf black swan}, {\sf car-turn}, {\sf disco ball}, {\sf hike}, and {\sf lucia}, which can be found in \href{https://lightbulb12294.github.io/hashing-nvd/#Reconstruction_1080p}{\sf Reconstruction (1080p)} section, with a video {\sf 1080p.mp4} showing simple comparisons between 1080p and 480p reconstructions.

\subsection*{A.2. Video Editing}

% \begin{table}
%     \begin{center}
%         \begin{tabular}{l|l}
%         video name (.mp4) & modification \\ \hline
%         bear\_styleTransfer & Style transfer is applied to both foreground and background.\\ \hline
%         discoball\_iccv & The wall on the right of the disco ball features a handwritten inscription of ``ICCV".\\ \hline
%         discoball\_relight & Periodically change the lighting reflecting off the disco ball.\\ \hline
%         hike\_darken & The background is darkened, and an additional light source is added on the right side.\\ \hline
%         hike\_darken+styleTransfer & Style transfer is applied to the background in addition to the previous modifications.\\ \hline
%         hike\_iccv+color & Handwritten ``ICCV" on the ground and recolor the shirt of the hiker.\\ \hline
%         kite-surf\_styleTransfer & Style transfer is applied to the background.\\ \hline
%         libby\_iccv & Handwritten ``ICCV" on the dog's body.\\ \hline
%         parkour\_headSpray & A contour of a man's head is spray-painted on the ground.\\ \hline
%         sunset\_batman & Batman logo hanging on the sky and an angry emoji floating on the river.
%         \end{tabular}
%     \end{center}
%     \caption{
%     \textbf{Explanation of modification of each video.}
%     All edited videos can be found in the ``editing'' directory.
%     \label{tab:editing}
%     }
% \end{table}

We showcase our video editing results, which can be found in \href{https://lightbulb12294.github.io/hashing-nvd/#Editing}{\sf Video Editing} section of our project page.

\subsection*{A.3. Camera Motion Manipulation}

% \begin{table}
%     \begin{center}
%         \begin{tabular}{l|l}
%         video name (.mp4) & camera motion \\ \hline
%         blackswan\_fixedCamera & Fixed on a specific frame.\\ \hline
%         car-turn\_fixedCamera & Fixed on a specific frame.\\ \hline
%         car-turn\_slowToFast & Follows the original motion track, but with different speed and acceleration.\\ \hline
%         hike\_circleCamera & Fixed and rotates around a specific frame, drawing a circle.\\ \hline
%         hike\_fixedCamera & Fixed on a specific frame.\\ \hline
%         hike\_slowToFast & Follows the original motion track, but with different speed and acceleration.\\ \hline
%         kite-surf\_fixedCamera & Fixed on a specific frame.
%         \end{tabular}
%     \end{center}
%     \caption{
%     \textbf{Explanation of camera motion of each video.}
%     All videos featuring camera motion manipulation can be found in the ``camera'' directory.
%     \label{tab:camera}
%     }
% \end{table}

We manipulate camera motion in different ways. The demonstrations can be found in \href{https://lightbulb12294.github.io/hashing-nvd/#Camera}{\sf Camera Motion Control} section of our project page.

\subsection*{A.4. Comparison with Previous Work}

% \begin{table}
%     \begin{center}
%     \begin{tabular}{c|cccc|ccc}
%     & & Training & Rendering & GPU & & PSNR & \\
%     Method & Resolution &  time &  speed (fps) &  memory & {\sf bear} & {\sf disco ball} & {\sf hike} \\ \hline
%     % PSNR 的計算都是把 input video scale，所以低解析度算出來的數值會比較好看一點
%     Deformable Sprites~\cite{YeLTKS22}& $213\times120$ & 10 minutes & 5  & 5 GB & 22.7 & 26.2 & 21.5 \\
%     Deformable Sprites~\cite{YeLTKS22} & $427\times240$ & 20 minutes & 1.6  & 12 GB & 23.6 & 26.4 & 22.0 \\
%     Layered Neural Atlases~\cite{KastenOWD21} & $768\times432$ & 5.5 hours & 0.5  & 3 GB & 27.3 & 29.0 & 25.2 \\
%     Ours & $768\times432$ & 40 minutes & 787  & 3 GB & 27.5 & 37.7 & 25.6 \\ \hline
%     Ours & $1920\times 1080$ & 40 minutes & 71  & 5 GB & 26.9 & 35.6 & 25.4
%     \end{tabular}
%     \end{center}
%     \caption{
%     \textbf{Comparison results.}
%     We report the PSNR for each video, along with the corresponding training time, rendering speed (frames per second), and GPU memory usage under different resolutions.
%     Our work achieves better results than prior work on the three videos, achieving faster rendering, lower GPU memory consumption, and higher resolution. Note that we measure the PSNR at the corresponding reconstructed resolution. Therefore the PSNR tends to favor the evaluation at low resolution, explaining why our 1080 reconstructions have slightly lower PSNR than our 480p reconstructions. 
%     \label{tab:comparison}
%     }
% \end{table}

We present a comparison of our reconstruction results with those obtained using Layered Neural Atlases~\cite{KastenOWD21} and Deformable Sprites~\cite{YeLTKS22}. The comparison results are included in \href{https://lightbulb12294.github.io/hashing-nvd/#Comparison}{\sf Compare with Previous Works} section of our project page.
We also provide a comparison of the alpha masks for the {\sf bear} video, as we find that Layered Neural Atlases tend to compensate for variations in lighting through the alpha mask, resulting in noisy and inaccurate object masks.

% We also report the PSNR values and various settings for every approach in Table~\ref{tab:comparison}. Our method demonstrates superior performance of PSNR, rendering speed, and GPU memory consumption compared to previous methods on all three videos while maintaining the same or better resolution. Note that we could not run Deformable Sprites in the $768\times432$ resolution, as Deformable Sprites failed to finish due to their GPU memory usage exceeding the limit of a single GPU with 24 GB memory.
% In contrast, our method can efficiently decompose higher-resolution videos into full-HD layers using affordable computation resources while maintaining comparable reconstruction results.
% Note that in order to compute the PSNR, we scale the input video into the reconstructed resolution. Therefore, the PSNR of our 1080p model might be slightly lower than that of the $768\times432$ model, as the PSNR of the 1080p model is calculated at the corresponding resolution and considered harder due to having more details to reconstruct than the low-resolution counterpart.

% camera ready 新增
\subsection*{A.5. Multiple Foreground Objects}\label{sec:multi_objects}

\begin{figure*}
    \centering
    \includegraphics[width=1.0\linewidth]{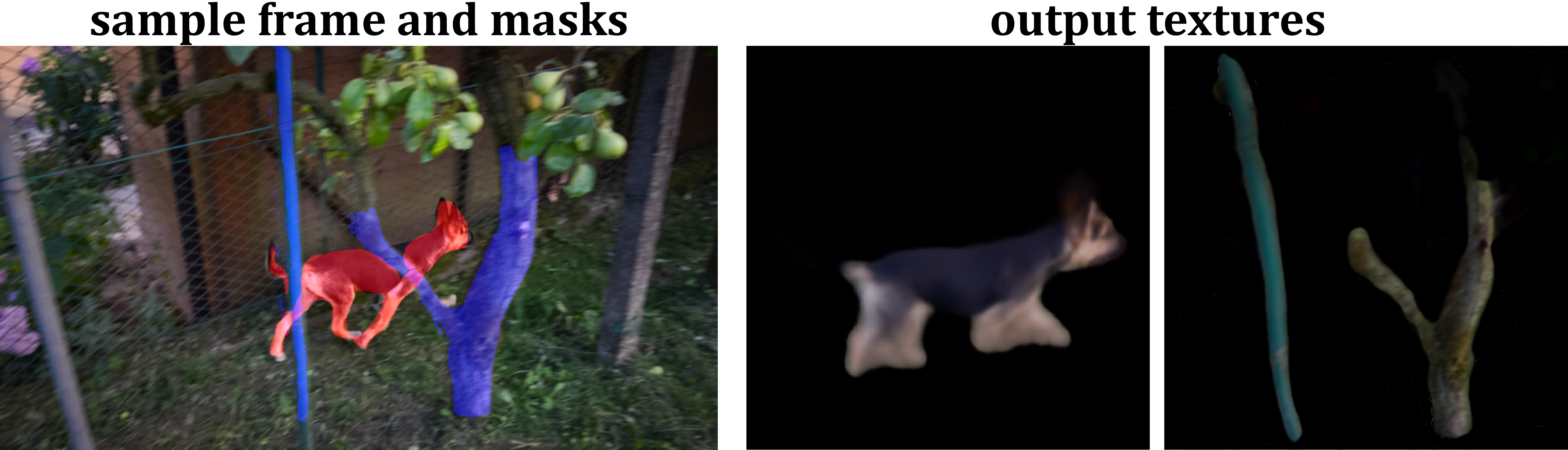}
    \caption{
    \textbf{Representation of multiple foreground objects.}
    We split two foreground objects for the video.
    Our method can handle multiple foreground objects, even when one object occludes another.
    }
    \label{fig:multiobject}
\end{figure*}

Our model can handle multiple foreground layers simultaneously.
In Fig.~\ref{fig:multiobject}, we present an example where the pipe and trunk are assigned to one foreground layer while the dog belongs to another.
The mask for the pipe and trunk is manually crafted for a single frame, while the remaining frames are generated using pre-computed optical flow, which may result in some inaccuracies in the generated masks.
Furthermore, the presence of occlusion between the dog and the pipe and trunk in certain frames poses a challenge for accurate decomposition.
Our method successfully decomposes the video into two foreground layers and a background, even in the presence of inaccurate masks and occlusion.

% camera ready 新增
\section*{B. Losses Inherited from Previous Work}

\subsection*{B.1. Reconstruction Loss}
We utilize the squared distance as our reconstruction loss, which is composed of two terms. The first term represents the distance between the ground truth and the predicted color, while the second term accounts for the image gradient.

\begin{equation}
\begin{split}
\mathcal{L}_{\text{RGB}}&=\lambda_{\text{RGB}}\left\lVert \hat{c}^{(p)}-\bar{c}^{(p)} \right\rVert^2_2 \,, \\
\mathcal{L}_{\text{Grad}}&=\lambda_{\text{Grad}}\left(\left\lVert \hat\nabla_x^{(p)}-\bar{\nabla}_x^{(p)} \right\rVert^2_2 + \left\lVert \hat\nabla_y^{(p)}-\bar{\nabla}_y^{(p)} \right\rVert^2_2\right) \,,
\end{split}
\end{equation}
where $\bar{c}^{(p)}$ and $\left( \bar{\nabla}^{(p)}_x, \bar{\nabla}^{(p)}_y \right)$ are the ground truth color and spatial derivatives of pixel $p$, and $\hat c^{(p)}$ and $\left( \hat{\nabla}^{(p)}_x, \hat{\nabla}^{(p)}_y \right)$ are our predictions, respectively.

\subsection*{B.2. Sparsity Loss}
To avoid the presence of duplicate foreground objects in the texture area, we incorporate a sparsity loss inspired by Layered Neural Atlases~\cite{KastenOWD21}.
For a given pixel $p$ in a video frame, if $p$ is mapped to the background, it is invisible for any other foreground objects.
Consequently, the color value of $p$ in the invisible area should be purely black, devoid of any foreground information. Hence, we incentivize the color of $p$ to be black through the sparsity loss:
\begin{equation}
\mathcal{L}_{\text{sparsity}}=\lambda_{\text{sparsity}}\sum_{i=0}^{N-1}\left\lVert (1-\alpha_i)c_i \right\rVert^2 \,.
\end{equation}

% \paragraph{Rigidity loss.}
% TODO: 欸恭喜真的沒用

\subsection*{B.3. Optical Flow Loss}
To ensure consistent mapping between points in the scene and corresponding points on the texture, we utilize a pre-trained optical flow model~\cite{TeedD20} and incorporate the optical flow loss from the previous approach~\cite{KastenOWD21}.
This loss encourages the predicted optical flow to align with the ground truth optical flow, enabling accurate mapping between the scene and texture coordinates.
In particular, for consecutive frames at time $i$ and $i+1$, we aim to achieve consistency in both the alpha and color values using optical flow as follows:
\begin{equation}
\begin{split}
\mathcal{L}_{\text{of-c}}&=\lambda_{\text{of-c}}\sum_{i=0}^N {\alpha_i^{(p)} \left\lVert \mathcal{M}_i(p)-\mathcal{M}_i(p') \right\rVert} \,,\\
\mathcal{L}_{\text{of-}\alpha}&=\lambda_{\text{of-}\alpha}\sum_{i=0}^N {\left| \alpha_i^{(p)} - \alpha_i^{(p')} \right|} \,,
\end{split}
\end{equation}
where $\mathcal{M}_i$ represents the mapping network of layer $i$, and $p'$ is the corresponding point of $p$ in either backward or forward direction.

\subsection*{B.4. Alpha Bootstrapping Loss}
Given that our model lacks prior knowledge about the objects present in the scene, we leverage the use of coarse masks obtained from Mask-RCNN~\cite{HeGDG20} as initial guidance for our model.
\begin{equation}
\mathcal{L}_{\text{bootstrap}}=\lambda_{\text{bootstrap}}\text{BCE}\left( m,\alpha \right) \,,
\end{equation}
where $m$ is the coarse mask.
The loss would be deactivated after a period of training.

\section*{C. Implementation Details}
% TODO: 一堆實作細節 (寫在 ppt 裡面)
% TODO: training 環境, learning rate, loss weight, 使用到的額外資源
% 可以提到 object mask 能用 deformable sprites 的方式來取代?
Our standard experimental setup involves videos comprising $50$ to $100$ frames with a resolution of $768\times432$.
We sample $10000$ points for each iteration to train our model and train it for a total of $50000$ iterations.
We use RAFT~\cite{TeedD20} to compute the optical flow and Mask R-CNN~\cite{He2017MaskR} to generate an initial coarse object mask for every frame.
The entire training process takes approximately 20--40 minutes which requires 3GB of GPU RAM and can process $71$ frames per second during inference for both edited and unedited videos using an NVIDIA RTX 3090 Ti GPU.

To speed up the inference process, we cache the UV coordinates and alpha masks for all frames, as well as the textures that are independent of time. This allows users to modify the textures and view the edited results in real-time without any delay.

We follow Layered Neural Atlases~\cite{KastenOWD21} to set the loss hyperparameters but divide the values by $1000$ to stabilize the training procedure. Additionally, we adopt alpha bootstrapping and pre-training stages to further improve our results.
We set $\lambda_{\mathcal{R}\mathrm{con}}=0.1$, $\lambda_{\mathcal{R}\mathrm{reg}}=0.5$, and $\lambda_{\alpha \mathrm{reg}}=0.1$ for the hyperparameters of loss terms.
As the video may include a large range of backgrounds, one might need to set the UV mapping scale to a lower value for different videos. In our experiments, we set the scale to $0.6$.
In our design, we prevent the gradient of the multiplicative residual from backpropagating to the mapping network; otherwise, it would lower the quality in the bootstrapping stage and is unstable, which might degrade the reconstruction quality.

We sample another batch of patches only from the edges in the video for computing the residual consistency loss.
Once an edge patch centered at $(x,y,t_1)$ is sampled, we randomly choose $15$ more frames as $t_2$ to calculate the residual consistency loss.
We then mask out all $(u,v,t_2)$ that are visible at time $t_2$. Note that the gradient is not backpropagated to $t_1$, since $t_1$ serves as the supervision for all $t_2$.
We may consider normalizing the raw data before computing the correlation, as in some cases, the computed standard deviation and covariance may be too small.

{\small
\bibliographystyle{ieee_fullname}
\bibliography{nvr}
}

\end{document}

%% file: macro.tex
\usepackage{bbm}
\usepackage{bm}
\usepackage{xspace}
\usepackage[utf8]{inputenc} % allow utf-8 input
\usepackage{url}            % simple URL typesetting
\usepackage{booktabs}       % professional-quality tables
\usepackage{amsfonts}       % blackboard math symbols
\usepackage{nicefrac}       % compact symbols for 1/2, etc.
\usepackage{microtype}      % microtypography
\usepackage{color}
\usepackage{algorithm}
\usepackage{algorithmic}
\usepackage{array}
\usepackage{multirow}
\usepackage{enumitem}
\usepackage{makecell}
\usepackage{gensymb}         % for degree
\usepackage{mathtools}
\usepackage{dblfloatfix}
\usepackage{pifont}
\usepackage[permil]{overpic}

\usepackage{titletoc,tocloft}

\usepackage{subcaption}

\makeatletter
\DeclareRobustCommand\onedot{\futurelet\@let@token\@onedot}
\def\@onedot{\ifx\@let@token.\else.\null\fi\xspace}

\def\eg{\emph{e.g}\onedot} 
\def\ie{\emph{i.e}\onedot}

\makeatother

\usepackage{xcolor}
\usepackage{ifthen}
\usepackage{textcomp}
\definecolor{red}{rgb}{1,0,0}
\definecolor{slateblue}{rgb}{0.7,0.35,0.9}
\definecolor{green}{rgb}{0,1,0}
\definecolor{mahogany}{rgb}{0.75, 0.25, 0.0}
\definecolor{purple}{rgb}{0.6, 0, 0.6}
\definecolor{darkpurple}{rgb}{0.3, 0, 0.3}
\definecolor{darkgreen}{rgb}{0, 0.4, 0}
\definecolor{frenchblue}{rgb}{0.0, 0.45, 0.73}
\definecolor{blue}{rgb}{0,0,1}
\definecolor{goldenrod}{rgb}{0.65, 0.45, 0.03}
\definecolor{gray}{rgb}{0.5,0.5,0.5}
\definecolor{gold}{rgb}{1.0, 0.874, 0}
\definecolor{silver}{rgb}{0.67,0.67,0.67}
\definecolor{brown}{rgb}{0.8, 0.678, 0.4}

\newboolean{revising}
\setboolean{revising}{true}
\ifthenelse{\boolean{revising}}
{
    \newcommand{\ignore}[1]{}
    \newcommand{\chengsun}[1]{\textcolor{darkpurple}{#1}}

} {
    \newcommand{\ignore}[1]{}
    \newcommand{\chengsun}[1]{}
}

\makeatletter
\newcommand\footnoteref[1]{\protected@xdef\@thefnmark{\ref{#1}}\@footnotemark}
\makeatother